# In-situ process monitoring for defect detection in wire-arc additive manufacturing: an agentic AI approach

Pallock Halder[1], Satyajit Mojumder[1*]

[1]School of Mechanical and Materials Engineering, Washington State University, Pullman, WA, USA

**Abstract:**

AI agents are being increasingly deployed across a wide range of real-world applications. In this paper, we propose an agentic AI framework for in-situ process monitoring for defect detection in wire-arc additive manufacturing (WAAM). The autonomous agent leverages a WAAM process monitoring dataset and a trained classification tool to build AI agents and uses a large language model (LLM) for in-situ process monitoring decision-making for defect detection. A processing agent is developed based on welder process signals, such as current and voltage, and a monitoring agent is developed based on acoustic data collected during the process. Both agents are tasked with identifying porosity defects from processing and monitoring signals, respectively. Ground truth X-ray computed tomography (XCT) data are used to develop classification tools for both the processing and monitoring agents. Furthermore, a multi-agent framework is demonstrated in which the processing and monitoring agents are orchestrated together for parallel decision-making on the given task of defect classification. Evaluation metrics are proposed to determine the efficacy of both individual agents, the combined single-agent, and the coordinated multi-agent system. The multi-agent configuration outperforms all individual-agent counterparts, achieving a decision accuracy of 91.6% and an $F_1$ score of 0.821 on decided runs, across 15 independent runs, and a reasoning quality score of 3.74 out of 5. These in-situ process monitoring agents hold significant potential for autonomous real-time process monitoring and control toward building qualified parts for WAAM and other additive manufacturing processes.

*Corresponding Author (Satyajit Mojumder, satyajit.mojumder@wsu.edu)

# 1. Introduction

Additive manufacturing (AM) has fundamentally transformed the production landscape by enabling the layer-by-layer fabrication of complex geometries with high material efficiency and design flexibility [1,2]. Among the diverse AM modalities, WAAM has garnered considerable attention owing to its high deposition rates, low equipment cost, and capacity to produce large-scale metallic components from a wide range of engineering alloys [3,4]. WAAM employs an electric arc as the heat source and a metallic wire as the feedstock, leveraging established gas metal arc welding (GMAW), gas tungsten arc welding (GTAW), or plasma arc welding (PAW) processes in a directed energy deposition (DED) configuration [5,6]. The technique has demonstrated significant promise in aerospace, maritime, energy, and construction sectors, where the demand for cost-effective production of large structural parts is acute [7,8]. Despite its advantages, WAAM is susceptible to a variety of process-induced defects that can compromise the structural integrity and mechanical performance of fabricated components. Porosity, lack of fusion, cracking, residual stress, and geometric deviations are among the most commonly reported defects in WAAM-produced parts [9,10]. Porosity has been identified as a critical concern, as it can reduce tensile strength and fatigue life, while cracks can lead to strength reductions [11,12]. These defects arise from complex interactions among process parameters, including arc current, voltage, wire feed rate, travel speed, and shielding gas composition, and the thermo-mechanical history experienced by the deposited material [13,14]. The quality and repeatability of WAAM parts are therefore highly sensitive to these parameters, necessitating advanced monitoring and adaptive control strategies to ensure part qualification [15,16].

In-situ process monitoring has emerged as an indispensable approach for real-time quality assurance in AM processes [17,18]. By acquiring and analyzing sensor data during the deposition process, in-situ monitoring enables the early detection of anomalies and process deviations before they propagate into critical defects [19,20]. For WAAM specifically, a diverse array of sensing modalities has been explored, encompassing electrical signal monitoring (current and voltage), acoustic emission sensing, thermal imaging, visual inspection, and spectroscopic analysis [21]. These sensing approaches can be broadly categorized into direct process signatures, such as arc current, arc voltage, and wire feed dynamics, and indirect indicators derived from the thermal field, acoustic emissions, and melt pool morphology [22,23]. Recent reviews have comprehensively surveyed the state of the art in WAAM monitoring technologies, identifying sensor fusion and multi-modal data integration as key research frontiers [24,25].

Electrical signal monitoring, particularly the analysis of welding current and voltage waveforms, constitutes one of the most widely adopted approaches for WAAM process assessment [26,27]. The rationale underlying this approach is that perturbations in arc stability, metal transfer mode, and deposition quality manifest as characteristic signatures in the time-domain and frequency-domain representations of current and voltage signals [28,29]. When both current and voltage exhibit regular periodic fluctuations, this is indicative of stable short-circuit droplet transfer, whereas sudden changes in these signals signify abnormal transfer modes and potential defect formation [30]. Time-frequency analysis methods, including wavelet decomposition and variational modal decomposition (VMD), have been employed to discriminate between normal and anomalous WAAM processes, revealing that voltage signals of defective weld beads exhibit distinctive fluctuations in their peak values [31,32]. Multi-source signal fusion strategies combining current, voltage, and sound signals with one-dimensional convolutional neural networks (1D-CNN) have demonstrated promising capabilities for identifying different defect states in WAAM [33].

Acoustic and audio-based monitoring represents another important sensing modality for in-situ defect detection in welding and AM [34,35]. The physical basis for acoustic monitoring lies in the fact that the arc welding process generates characteristic audio signatures that are correlated with process stability, metal transfer dynamics, and defect formation mechanisms [36,37]. Acoustic emission sensors and microphone-based systems have been deployed to capture both airborne acoustic signals and structure-borne elastic waves during deposition, enabling the detection of porosity, cracking, and arc instability [38,39]. Recent advances have demonstrated that audio models developed specifically for detecting porosity, overlap, and crater cracks can achieve detection latencies as low as 42.7 milliseconds, making them suitable for real-time monitoring applications [40]. Frequency-domain analysis of acoustic time series has proven effective for identifying anomalies such as porosity, spatter, arc shifts, and contamination, with machine learning classifiers achieving accuracies exceeding 95% on extracted spectral features [41,42]. Unsupervised approaches combining audio and video data have also been explored for welding defect detection, offering the advantage of reducing reliance on labeled training data [43,44].

Establishing reliable ground truth for defect characterization is essential for training and validating data-driven monitoring systems. X-ray computed tomography (XCT) has emerged as the gold standard for non-destructive volumetric characterization of internal defects in additively manufactured components [45,46]. XCT enables three-dimensional reconstruction of porosity distributions, pore size and shape analysis, and deformation mapping with micrometer-scale resolution [47,48]. In the context of WAAM, XCT data provide invaluable ground truth labels for correlating process signatures with internal defect states, thereby enabling the development of supervised classification models [49,50]. Recent studies have investigated how different segmentation methodologies applied to XCT data influence machine learning performance for porosity detection, highlighting the importance of standardized protocols for ground truth generation [51]. Deep learning-based approaches for porosity segmentation in XCT scans of metal AM specimens have achieved notable success, with frameworks leveraging high-quality reference datasets established through mechanical polishing serial sectioning to benchmark XCT-derived measurements [52,53].

Machine learning (ML) and deep learning (DL) methodologies have been extensively applied to defect detection and classification in AM, driven by the increasing availability of sensor data and computational resources [54,55]. CNNs have demonstrated efficacy for image-based defect classification, achieving $F_1$ scores up to 0.96 across multiple defect categories, including porosity, lack of fusion, cracking, and spattering [56,57]. YOLOv8-based architectures have achieved enhanced defect detection accuracy of up to 96% for cracks and porosity in metal AM datasets [58], while hybrid CNN-LSTM models trained on multi-sensor data from acoustic emission, infrared, and visible sensors have attained classification accuracies ranging from 95.9% to 100% across lack of fusion, conduction mode, and keyhole regimes [59,60]. In the WAAM domain specifically, deep learning-based image segmentation methods have been proposed for online metallic surface defect detection [43], for instance, YOLO-attention models have obtained a mean average precision of 94.5% with processing rates exceeding 42 frames per second [61]. Wavelet scattering networks combined with sparse principal component analysis (sPCA) have been applied to acoustic and current signals for predicting porosity in WAAM [62]. Notwithstanding these advances, significant challenges remain in developing models that generalize across different materials, process parameters, and equipment configurations [63,64].

The emergence of LLMs has catalyzed a paradigm shift across numerous scientific and engineering domains [65,66]. Built upon the transformer architecture and trained on massive text corpora, LLMs such as GPT-4, Claude, LLaMA, and Gemini have demonstrated remarkable capabilities in natural language understanding, code generation, reasoning, and in-context learning [67,68]. The adaptability of LLMs to new domains through techniques such as chain-of-thought (CoT) prompting, retrieval-augmented generation (RAG), and few-shot learning has enabled their deployment in diverse applications spanning chemistry, materials science, biomedical research, and engineering design [69–71]. In the manufacturing context, LLMs have been leveraged for contextual querying of AM knowledge bases [72], G-code comprehension and manipulation [73], quality management assistance [74], and scheduling optimization in human-robot collaborative manufacturing systems [75]. Recent comprehensive reviews have surveyed the integration of LLMs into digital manufacturing, identifying their potential to serve as cognitive assistants that govern and control manufacturing processes by interpreting sensor data and coordinating decision-making across digital system levels [76,77].

A particularly transformative development in the LLM landscape is the emergence of autonomous agentic artificial intelligence (AI) systems that combine LLMs with tool-calling capabilities, memory, planning, and iterative reasoning to accomplish complex, multi-step tasks [78,79]. The ReAct (Reasoning and Acting) framework, introduced by Yao et al. [80], established a foundational paradigm for interleaving reasoning traces with task-specific actions, enabling LLM agents to interface with external tools, databases, and computational environments. Subsequent frameworks including LangChain, LangGraph, AutoGen, and CrewAI have provided increasingly sophisticated orchestration infrastructures for building single-agent and multi-agent systems [81–83]. LangGraph, in particular, has emerged as a prominent agent runtime that supports diverse control flows such as single-agent, multi-agent, and hierarchical architectures through a graph-based representation where agents are modeled as nodes, their interactions as edges, and coordination is managed through shared state [84]. Tool-augmented agents extend the cognitive capabilities of LLMs by enabling function calling to external APIs, computational tools, and domain-specific software, forming the operational backbone of contemporary agentic AI systems [85,86]. Multi-agent systems represent a natural extension of single-agent architectures, wherein multiple specialized agents collaborate, communicate, and coordinate to solve complex problems that exceed the capabilities of any individual agent [87,88]. In the LLM-based multi-agent paradigm, each agent is typically endowed with a distinct prompt, tool suite, and domain expertise, enabling modular task decomposition and sequential or parallel decision-making [89,90]. Surveys on LLM-based multi-agent systems have identified three primary architectural components: perception, brain (reasoning and planning), and action modules, that collectively enable autonomous operation in dynamic environments [91]. The AtomAgents platform exemplifies this approach in materials science, synergizing multiple AI agents with expertise in knowledge retrieval, multi-modal data integration, and physics-based simulations for autonomous alloy design [92]. Similarly, MatSciAgent demonstrates a modular multi-agent framework for computational materials science tasks, including data retrieval, crystal structure generation, and molecular dynamics simulation [93]. In the manufacturing domain, agentic AI frameworks have been proposed that integrate LLM-based agents with unified data-model-knowledge architectures for goal-directed manufacturing intelligence [94,95].

The application of LLM-based agents to AM process monitoring and control represents an emerging frontier with significant potential. Jadhav et al. [96] introduced LLM-3D Print, a framework employing LLMs as autonomous controllers for material extrusion 3D printing, demonstrating that LLM-based agents can identify defects such as inconsistent extrusion, stringing, warping, and poor layer adhesion, determine root causes, and execute corrective actions without human intervention. Pak et al. [97] developed an agentic system for AM alloy evaluation, wherein LLM-enabled agents dispatch tool calls via the Model Context Protocol (MCP) to perform thermophysical property calculations and lack-of-fusion process map generation. Chandrasekhar et al. [72] proposed AMGPT, a specialized LLM text generator for contextual querying in additive manufacturing, leveraging retrieval-augmented generation with a curated corpus of AM literature. Merrill et al. [98] demonstrated LLM-Drone, integrating LLMs with drone-based AM for semantic planning and real-time error correction. Guo et al. [99] introduced LLM-empowered computer-aided engineering (CAE) agents capable of autonomously planning, executing, and adapting simulation workflows. These pioneering works demonstrate the viability of LLM-based agents for manufacturing tasks, yet their application to in-situ process monitoring, particularly for WAAM, remains largely unexplored.

Despite the considerable progress in both in-situ monitoring for WAAM and agentic AI systems, a significant research gap persists at their intersection. Existing monitoring approaches predominantly rely on dedicated ML/DL models that, while effective within their trained domains, lack the reasoning flexibility, interpretability, and adaptive decision-making capabilities inherent to LLM-based agents [100,101]. Conversely, the emerging body of work on LLM agents for manufacturing has primarily focused on process planning, parameter optimization, and post-hoc quality assessment rather than real-time in-situ monitoring with multi-modal sensor data[102,103]. The potential for autonomous agents to ingest heterogeneous process signals (electrical, acoustic, and visual), reason about their implications for part quality, and coordinate multi-agent decision-making for defect classification has not been systematically investigated [104].

In this paper, we propose an agentic AI framework for in-situ process monitoring of WAAM, addressing this gap at the confluence of WAAM quality assurance and autonomous AI systems. The proposed framework comprises a processing agent that analyzes welding current and voltage signals and a monitoring agent that processes acoustic data collected during deposition. Using a LangGraph-based workflow with an LLM, the framework is implemented and evaluated as individual single agent, combined single-agent, and multi-agent architectures. A ground truth dataset derived from XCT is employed to develop classification tools for porosity defect identification, which are integrated as callable tools within the agent framework. Evaluation metrics such as decision accuracy, monitoring utility score, and reasoning quality are proposed to assess the efficacy of both individual agents and the coordinated multi-agent system. To the best of our knowledge, this work represents the first application of agentic AI with LLM-orchestrated multi-agent frameworks for in-situ process monitoring in WAAM, opening new avenues for autonomous, real-time quality assurance in metal AM.

## 2. Methodology

### 2.1 Agentic framework overview

This section presents the proposed agentic AI framework for in-situ process monitoring for porosity detection in the WAAM printed part. The overall architecture follows a modular pipeline, that can be swapped with other tools and models, consists of three functional layers: (i) a signal acquisition layer that captures multi-modal process and monitoring data, (ii) a machine learning tool layer that performs individual signal-level classification along with other functional tools for preprocessing, retrieval of information and result aggregation, and (iii) an agent orchestration layer in which LLMs reason over tool outputs to produce a final monitoring decision. The framework is designed to be modular; the classification tools and the underlying LLM are independently replaceable, which allows future upgrades to either component without restructuring the overall system.

Two distinct signal modalities are employed to capture processing and monitoring aspects of the WAAM process. The first modality consists of process signals (welding current and voltage), which are recorded by the data acquisition system. The second modality consists of monitoring signals captured by a microphone positioned near the welding torch, encoding the audio signatures of the welding process. This audio data is later processed in Short-Time Fourier Transforms (STFT) spectrograms using relevant tools by the agents for further classification and decision making. These two modalities are processed by dedicated trained classification tools: a 1D-CNN for processing signals and a 2D-CNN for monitoring signals. Each tool accepts the raw signal data for a given layer's given track, executes the model inference, and returns a structured text output comprising a classification label (Normal or Defective) and an associated confidence score to the LLM agent. By defective, it means a porosity formation in the track, which is available from ground truth XCT analysis of the printed part. More details on the WAAM dataset used to develop this agent are provided in Section 2.2.

The LLM agent receives these text-based tool outputs and engages in structured reasoning to produce a monitoring decision. Critically, the agent does not operate directly on the raw sensor data; rather, it functions as a reasoning layer that interprets, synthesizes, and adjudicates the outputs of specialized ML tools. This design philosophy aligns with the tool-augmented agent paradigm, wherein the LLM's strength in natural language reasoning is complemented by domain-specific computational tools that handle signal-level feature extraction and classification [105,106].

The framework supports four operational configurations for decision-making. In the individual single-agent configuration, either the processing agent (equipped with the 1D-CNN tool for processing signals) or the monitoring agent (equipped with the 2D-CNN tool for monitoring signals) operates independently, producing a decision from a single modality. In the combined single-agent configuration, a single LLM agent has access to both tools and reasons over both modalities in one inference pass. In the multi-agent configuration, three specialized LLM agents: a processing agent, a monitoring agent, and an orchestrating agent, collaborate through a hierarchical workflow to produce a coordinated decision. **Figure 1** presents the system-level architecture of the proposed framework.

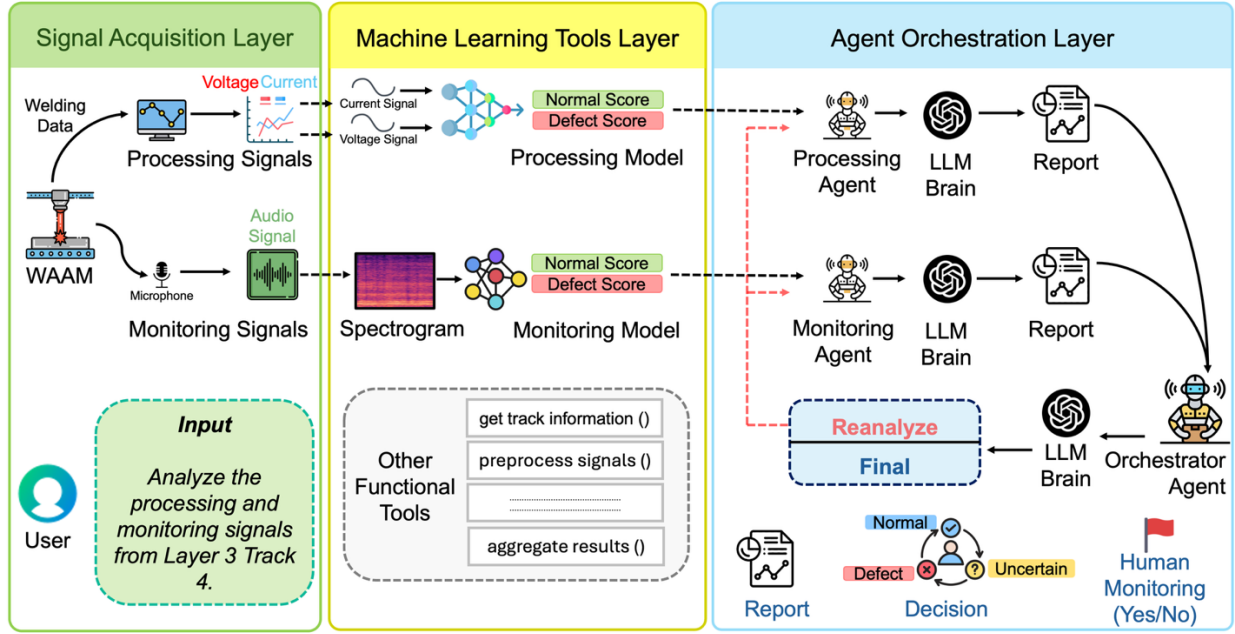

**Figure 1.** System overview of the proposed agentic AI framework for in-situ WAAM process monitoring. The architecture comprises three functional layers: (a) a signal acquisition layer capturing processing signals (welding current and voltage) and monitoring signals (acoustic data), (b) a machine learning tool layer with a 1D-CNN for processing signal classification and a 2D-CNN for monitoring signal (audio spectrogram) classification together with functional tools for signal retrieval, preprocessing, and result aggregation, and (c) an agent orchestration layer in which specialist LLM agents receive structured tool outputs and reason independently over each modality before an orchestrator agent synthesizes a final track-level monitoring decision. The orchestrator may trigger re-analysis of either modality when evidence is ambiguous, as indicated by the dashed feedback pathway. The modular design permits independent replacement of classification tools and the underlying LLM.

**2.2 Dataset**

The WAAM dataset used in this study was originally generated by Orlyanchik et al. [107] in the Oak Ridge National Laboratory (ORNL) manufacturing demonstration facility. This publicly available dataset contains in-situ and ex-situ data collected during the WAAM deposition of a rectangular mild steel block using a Tormach ZA6 robotic arm equipped with a Lincoln Electric Power Wave R450 welder and AutoDrive 4R220 wire feeder. Deposition was carried out in rapid X welding mode with 95/5 Ar-$CO_2$ shielding gas, producing a 15-layer part consisting of six unidirectional tracks per layer. All in-situ process data were acquired through the Robot Operating System (ROS), ensuring synchronized timestamps across all sensors. The recorded data included robot position (16.7 Hz), welder current and voltage (100 Hz), wire feed speed and feeder current (100 Hz), and acoustic emission (50 kHz) captured by an Ultramic UM200k microphone. After deposition, the part was characterized using XCT with a Metrotom 220 kV scanner, achieving a voxel size of 25.92 µm. The XCT volumes were reconstructed with beam-hardening correction and segmented using a 2.5D U-Net model to identify internal flaws, which were spatially registered to the robot coordinate system using the STL reference and iterative closest point alignment. The dataset is organized in HDF5 format, containing synchronized groups for audio, welder, wire feeder, and robot position data, as well as XCT data and the part geometry, providing a fully aligned in-situ and ex-situ record for defect analysis in WA-DED. More details of the experimental setup can be found in the source paper [107].

### 2.3 Classification tools

The classification tools employed in this framework are pre-trained CNN models that serve as callable inference endpoints for the LLM agents. Both models were trained on the ORNL WAAM dataset, comprising approximately 90 deposited tracks over 15 layers with porosity labels derived from XCT ground truth data. A detailed description of the dataset, training procedures, and standalone model performance is provided in the following sections and the **Appendix A**; the present section provides a summary of each tool's architecture and interface specification.

#### 2.3.1 Processing signal tool

The processing signal tool is built upon a 1D-CNN architecture, consisting of three convolution layers, designed to classify fixed-length time windows of processing (welding current and voltage) signals. Each input consists of a two-channel time series data, welding current and voltage, sampled at 100 Hz and windowed into a 2-second time window (200 samples per channel) with a 0.5-second stride. The convolutional backbone applies successive Conv1d layers with kernel sizes of 7, 5, and 3, each followed by Batch Normalization, ReLU activation, and channel dropout (0.2). Max pooling is applied after the first two layers to progressively reduce temporal resolution, while the third layer preserves spatial extent. A global average pooling operation then reduces the temporal dimension, producing a 64-dimensional feature vector that is passed through two fully connected layers (64→32→2) with ReLU activations and a final dropout rate of 0.3. The model is trained with Focal Loss ($\alpha = 0.90$, $\gamma = 3.0$) to address the severe class imbalance (~4.3% positive or defect windows), using the AdamW optimizer with a linear warmup schedule over 20 epochs and early stopping (patience = 50). Upon execution, the tool returns a structured text string to the calling agent in the format: {label: Normal/Defective, confidence: 0.XX}. Window-level validation against XCT ground truth yielded an area under the receiver operating characteristic curve (AUC) of 0.925 with perfect recall (zero false negatives) on the validation tracks. On the unseen test tracks, AUC was 0.677, with the model maintaining perfect recall at the cost of a high false positive rate. This is a desired tool characteristic that motivates its integration into a multi-modal reasoning framework rather than standalone deployment.

### 2.3.2 Monitoring signal tool

The monitoring signal tool is built upon a four-layer 2D-CNN that operates on STFT spectrograms derived from monitoring signals recorded during layer-wise track deposition. Raw audio is captured at 49,932 Hz and segmented into 2-second windows with a 0.5-second stride. Each window is transformed by STFT ($n_{fft}$ = 1024, hop length = 256) and converted to a decibel-scaled magnitude spectrogram, which is then min-max normalized and bilinearly resized to a fixed 128 × 128 grid.

The convolutional backbone consists of four Conv2d layers with progressively increasing channel widths (1→16→32→64→128), each followed by Batch Normalization and ReLU activation. Max pooling is applied after the first three layers to reduce spatial resolution, while the fourth layer preserves spatial extent at 16 × 16. Global average pooling then reduces the spatial dimensions into a 128-dimensional feature vector, which is passed through three fully connected layers (128→64→32→2) with ReLU activations and dropout regularization (0.6 and 0.5, respectively). The model is trained with Focal Loss ($\alpha$ = 0.80, $\gamma$ = 2.5) using the AdamW optimizer with a ReduceLROnPlateau learning rate schedule. Validated against XCT ground truth on tracks unseen during training, the audio tool achieved a validation AUC of 0.938 and a test AUC of 0.905, with perfect recall on the test set (zero false negatives at the selected operating threshold). The inference outputs are returned as a structured text string — {label: Normal/Defective, confidence: 0.XX} — for the agent layer.

### 2.4 LangGraph orchestration and tool integration

In this work, LangGraph is used to implement all four agentic configurations evaluated: the processing agent (welding/ processing signals only), the monitoring agent (audio/monitoring signals only), the combined single-agent, and the multi-agent system. The first three configurations follow a common pipeline structure consisting of an agent node, a tool execution node, a detection extraction node, a reflection node, and a report generation node in a consistent topology, differing only in which tools are registered and which system prompt is supplied. The multi-agent configuration extends this design with dedicated two parallel specialist nodes for the processing and monitoring agents, each operating on private message threads to prevent cross-contamination of tool outputs, plus an orchestrator node that synthesizes the two specialist reports into a final decision, as shown in **Figure 2.**

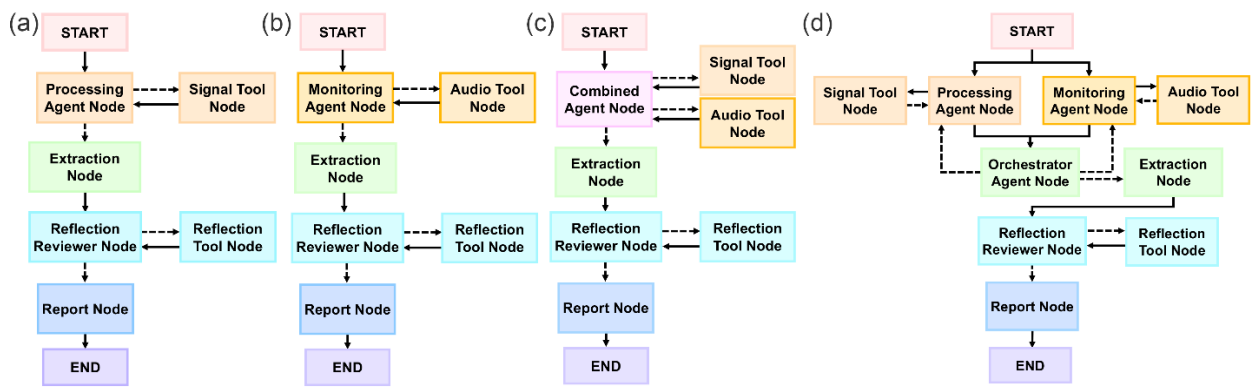

Figure 2. LangGraph workflow graphs for the four agentic orchestration configurations. (a) Processing-agent architecture: a single processing agent invokes the signal tool and performs reasoning on process signals, followed by extraction, reflection-based review, and report generation. (b) Monitoring-agent architecture: a single monitoring agent invokes the audio tool to analyze acoustic data, followed by extraction, reflection, and reporting. (c) Combined single-agent architecture: a unified agent invokes both processing and monitoring tools within a single reasoning step, integrating multimodal information before passing through extraction, reflection, and reporting stages. (d) Multi-agent architecture: the processing agent and monitoring agent operate as parallel specialized nodes, each invoking their respective tool and generating modality-specific outputs; an orchestrator agent synthesizes these outputs before extraction, reflection, and final reporting. Dashed arrows indicate optional feedback loops where agents may re-invoke tools or request re-analysis before proceeding. Agent and reviewer nodes correspond to LLM inference calls, while tool nodes denote external model executions (e.g., CNN-based process signal and monitoring signal analysis), and extraction/report nodes represent deterministic processing steps.

Tool integration follows the function-calling paradigm. The trained CNN tools are wrapped as callable Python functions and registered as LangChain tools within the graph. When the LLM emits a structured tool call, the LangGraph runtime intercepts it, dispatches it to the corresponding inference function, and returns the result as a formatted text string to the agent's message context. This text-based interface is deliberate: it allows the LLM to reason over CNN outputs as per-window defect probabilities and summary statistics without requiring direct access to model weights or tensors, and it preserves the modularity of the system by decoupling the ML layer from the reasoning layer entirely.

Prompt engineering constitutes a critical and non-trivial component of the agent design. Each agent is initialized with a system prompt that defines its analytical role, specifies the mandatory tool invocation sequence, and encodes modality-specific numerical reasoning rules calibrated to the known output behavior of each CNN. This calibration was necessary because both models produce probabilities in a compressed range (approximately 0.35–0.50 for the monitoring tool and 0.47-0.52 for the processing tool), meaning that naive threshold-based reasoning (e.g., flagging any window above 0.70 as defective) would never fire. The prompts instead instruct each agent to reason from intra-track statistics: computing the peak probability, minimum probability, mean probability, elevation (peak - mean), and contrast (peak / min) across all windows of a given track and using the contrast ratio as the primary discriminating metric. Each agent outputs key statistics, cluster location, a decision label, and a summary in a fixed format so downstream nodes can parse the results consistently.

State management is handled by LangGraph's built-in state persistence, which maintains the full conversational history, including tool calls, tool responses, and intermediate reasoning, across all nodes within a single workflow execution. In the multi-agent configuration, specialist agents additionally maintain private message threads, ensuring that the monitoring agent's tool outputs do not appear in the processing agent's context and vice versa. The LLM used across all agent nodes is GPT-4o-mini, with inference temperature set to 0 to ensure deterministic and reproducible outputs across evaluation runs.

### 2.5 Agentic orchestration configurations
### 2.5.1 Individual single modality-agent architecture

Two baseline configurations are evaluated in which a single agent operates with access to only one CNN tool. The processing agent is instantiated with access exclusively to the 1D-CNN process signal tool, while the monitoring agent is instantiated with access exclusively to the 2D-CNN monitoring signal tool. In each case, the agent executes the full preprocessing and inference pipeline for its respective modality, interprets the resulting per-window probability distribution described in Section 2.4, and produces a structured monitoring decision (Normal, Defect or Uncertain) followed by a reflection pass. These configurations serve as modality-specific baselines, enabling independent assessment of the discriminative contribution of each signal source and providing a reference against which the benefit of multi-modal fusion can be quantified.

### 2.5.2 Combined single-agent architecture

In the combined single-agent configuration, a single LLM agent is instantiated with access to both the processing signal tool and the monitoring signal tool. The agent is prompted to invoke both tools within the same reasoning loop by running the processing and monitoring pipeline before synthesizing the evidence from both modalities into a unified monitoring decision. Unlike the individual agents, this configuration requires the agent to interpret, compare, and reconcile outputs from two distinct CNN models in a single reasoning pass, without any modality-specific deliberation or intermediate report structure. The combined agent then passes its decision to the shared reflection node for review before report generation. This architecture is computationally leaner than the multi-agent configuration described in Section 2.5.3, as it requires only one LLM context rather than multiple independent instances. However, the reasoning burden is concentrated in a single pass over all tool outputs, which may limit the depth of modality-specific analysis, particularly given the compressed and asymmetric probability ranges of the two CNN models, which benefit from separate, calibrated interpretation.

### 2.5.3 Multi-agent architecture

The multi-agent configuration decomposes the monitoring task across three specialist LLM agents organized in a hierarchical pipeline: a processing agent, a monitoring agent, and an orchestrating agent. Each agent is instantiated as an independent LLM instance with a dedicated system prompt and access only to its designated tools, enforcing strict modality-specific specialization at both reasoning and tool-access levels.

Execution begins in parallel with the processing and the monitoring agent. The processing agent runs the inference pipeline as follows: preprocessing the processing (welding current and voltage) signals, invoking the 1D-CNN tool, and aggregating the per-window probability distribution before producing a structured process report containing its decision, key statistics (peak probability, elevation, contrast), cluster localization, and an interpretive summary. The monitoring agent executes analogously for the acoustic modality, running the full audio preprocessing and 2D-CNN inference pipeline and producing a corresponding monitoring report. Critically, each specialist agent operates on a private message thread, meaning that neither agent sees the other's tool outputs or intermediate reasoning during its own analysis. This isolation ensures that each modality is interpreted independently before any cross-modal synthesis occurs.

Once both specialist reports are available, the orchestrator receives them as structured text inputs and applies a fixed evidence hierarchy to produce a final monitoring decision. Given the monitoring model's superior discriminative performance (test AUC = 0.905 versus 0.677 for the process model), the orchestrator is explicitly prompted to treat monitoring evidence as the primary decision basis and processing evidence as secondary corroborating context. If the orchestrator determines that either specialist report is insufficient or ambiguous, it may request re-analysis of that modality by routing execution back to the relevant agent with a cleared message thread, which is subject to a hard cap of two reanalysis requests per track to prevent runaway loops. Following the orchestrator's decision, a fourth LLM instance serving as the reflection reviewer critically examines the decision, may invoke any tool from either modality for additional verification, and finalizes the report. In total, the multi-agent configuration instantiates four independent LLM instances, each with a distinct role and prompt, resulting in greater computational cost than the single-agent configurations but enabling deeper, modality-specific deliberation at each stage of the decision pipeline.

## 2.6 Evaluation strategy

Evaluating LLM-based agentic systems requires metrics that go beyond conventional classification accuracy, as the value proposition of agentic AI lies not only in the correctness of the final decision but also in the quality and reliability of the reasoning process. Accordingly, the evaluation strategy comprises three complementary dimensions: decision accuracy, monitoring utility, and reasoning quality. A qualitative assessment of agentic behavior through representative case studies is presented separately in the Results section.

### 2.6.1 Decision accuracy

Decision accuracy quantifies the degree to which the agent's final classification (Normal or Defective) agrees with the XCT-derived ground truth labels. For each test track $i$, the agent's predicted label $\hat{y}_i$ is compared against the ground truth label $y_i$. Standard binary classification metrics are computed, including overall accuracy, precision, recall, and $F_1$ score, defined as follows:

$$Accuracy = (TP + TN) / (TP + TN + FP + FN) \qquad (1)$$

$$Precision = TP / (TP + FP) \qquad (2)$$

$$Recall = TP / (TP + FN) \qquad (3)$$

$$F_1 = 2 \times (Precision \times Recall) / (Precision + Recall) \qquad (4)$$

where TP, TN, FP, and FN denote true positives, true negatives, false positives, and false negatives, respectively, with the Defective (porosity-present) class treated as the positive class. These metrics are reported for the overall test set as well as separately for normal and defective track subsets. To account for the stochastic nature of LLM-based agents, each test track is evaluated over multiple independent runs. Specifically, each of the nine test tracks is processed 15 times, resulting in a total of 135 evaluation instances per agent configuration. In each run, the agent operates without memory of previous interactions, ensuring that all predictions are generated independently. This repeated evaluation allows for a more robust assessment of agent performance by capturing variability in reasoning and decision-making and enables the computation of statistically stable performance metrics. Results are compared across all agent configurations: processing agent, monitoring agent, combined single-agent, and multi-agent.

### 2.6.2 Monitoring utility

Since all agent configurations exhibited deterministic behavior across repeated runs, decision consistency was not informative for distinguishing performance. Instead, we evaluate agents based on decision quality and operational usefulness, reflecting their role in real-world monitoring systems.

Each agent produces a structured output for every track, including a categorical decision (Normal, Defect, or Uncertain) and a monitoring flag indicating whether the track should be reviewed by a human. The Monitoring Utility Score is used to evaluate the usefulness of agent outputs in a monitoring context. A utility-based score is defined based on the correctness of the monitoring flag:

$$S_m = \begin{cases} 1, & if\ defect + flagged\ or\ normal + not\ flagged \\ 0, & otherwise \end{cases} \qquad (5)$$

$$Monitoring\ Score\ (MS) = \frac{1}{M}\sum_{i=1}^{M} S_m^{(i)} \qquad (6)$$

where *M* is the total number of test tracks. Results are additionally reported separately for normal and defective tracks to capture class-dependent behavior.

The monitoring score assesses its practical usefulness in a deployment setting. This metric provides a more comprehensive evaluation than standard classification measures, particularly for agent-based systems that incorporate uncertainty and decision support.

### 2.6.3 Reasoning quality

Beyond classification correctness and consistency, the quality of the agent's reasoning trace is evaluated to assess whether the agentic framework produces interpretable, logically sound, and factually grounded explanations. An automated evaluation protocol is employed using an independent LLM judge, specifically a separate GPT-4o-mini instance that does not participate in the monitoring task, to score each reasoning trace along three dimensions, each rated on a 1 to 5 Likert scale:

- **Factual grounding (FG):** Does the reasoning trace accurately reflect the tool outputs it received? A score of 5 indicates perfect fidelity to the reported labels and confidence scores with no fabricated or hallucinated information; a score of 1 indicates substantial misrepresentation of tool outputs.
- **Logical coherence (LC):** Is the reasoning logically structured, with conclusions that follow from the stated premises? A score of 5 indicates a clear, well-organized argument with explicit causal links between evidence and decision; a score of 1 indicates disjointed or self-contradictory reasoning.
- **Modality-specific insight (MSI):** Does the reasoning demonstrate understanding of what each signal modality reveals about the process? A score of 5 indicates a nuanced interpretation that distinguishes between the information content of processing signals versus monitoring signals; a score of 1 indicates generic reasoning that treats all tool outputs interchangeably.

The composite reasoning quality (RQ) score for each reasoning trace is computed as the arithmetic mean of the three dimensions:

$$RQ = (FG + LC + MSI) / 3 \tag{7}$$

The LLM judge is provided with the tool outputs, the agent's reasoning trace, the ground truth label, and a detailed scoring rubric as presented in **Appendix C**. Ground truth labels are withheld from the judge during scoring to ensure that reasoning quality is assessed independently of whether the final decision was correct, i.e., a well-reasoned but incorrect decision should score higher than a correct decision arrived at through poorly grounded logic. The mean RQ score is reported across all test tracks for each agent configuration, enabling quantitative comparison of the interpretability and reasoning depth afforded by the agents. This evaluation protocol follows established practices in the LLM evaluation literature, where LLM-as-judge approaches have demonstrated strong correlation with human assessments for structured reasoning tasks [108]. A known limitation of this approach is the potential for self-consistency bias: since the judge model (GPT-4o-mini) is the same model family as the agents being evaluated, it may systematically favor reasoning styles consistent with its own generation patterns. This limitation is acknowledged and should be considered in the interpretation of differences in reasoning quality scores across configurations.

## 3. Results

This section presents the quantitative and qualitative performance of individual processing and monitoring agents, a combined single-agent and multi-agent system for in-situ monitoring of WAAM. Performance is evaluated across classification tool performance, decision accuracy, monitoring utility, reasoning quality, and agentic behavior, including illustrative case studies of agentic behavior.

### 3.1 Classification tool performance

**Table 1** reports the standalone performance of both CNN tools on the ORNL test tracks. In many studies, the test set is rebalanced to contain equal numbers of defective and clean samples, which increases precision and $F_1$ scores. Here, the test set holds a percentage of 4.2% defective windows so that the reported metrics reflect what the model would encounter in actual production. The 1D-CNN scores 0.540 accuracy and 0.084 precision; the 2D-CNN reaches 0.743 and 0.141, respectively. Though these numbers being low, they portray the imbalance in the data distribution rather than the models. With defective windows making up only 1 in 24 observations, a classifier that blindly labels every window as clean would score 95.8% accuracy while catching zero defects. Conventional classification metrics such as accuracy, precision, and $F_1$ are not well-suited for evaluating models on heavily imbalanced manufacturing defect datasets, a limitation explicitly documented in the literature [109]. In contrast, the area under the receiver operating characteristic curve (AUC-ROC) and the precision recall area under the curve (PR-AUC) are widely recognized as more appropriate, as they are threshold independent and more robust to class imbalance.

From **Table 1**, both tools achieve a recall of 1.0 on the test set, meaning not a single defective window was missed. The 2D-CNN audio tool achieves an AUC-ROC of 0.905, which reflects a strong ability to rank defective windows above clean ones across all decision thresholds. The 1D-CNN processing tool reaches an AUC-ROC of 0.677, because the welding current and voltage sampled at 100 Hz carry considerably less defect-specific information than the acoustic emission captured at nearly 50 kHz. Therefore, any improvement observed in the multi-agent setting comes from how the tools are orchestrated together, not from the tools themselves being particularly strong.

**Table 1.** Standalone classification performance of the 1D-CNN (processing signal tool) and 2D-CNN (monitoring signal tool) on the held-out ORNL test tracks. Metrics computed with the defect class as positive.

| Tool | Accuracy | Precision | Recall | $F_1$ |
| --- | --- | --- | --- | --- |
| 1D-CNN (Processing) | 0.5402 | 0.0840 | 1.0000 | 0.1550 |
| 2D-CNN (Monitoring) | 0.7433 | 0.1410 | 1.0000 | 0.2472 |

### 3.2 Decision accuracy

This subsection compares the classification accuracy of four agent configurations (processing agent, monitoring agent, combined single-agent, and multi-agent) against XCT ground truth labels. Results are reported in terms of overall accuracy, precision, recall, and $F_1$ score, and are broken down by class (Normal/ Defective).

**Figure 3** presents confusion matrices for four agent configurations. The processing agent exhibits poor discriminative capability, achieving zero recall across all test runs, with every defective track classified as either Normal or Uncertain and no true positives detected. The large validation-to-test generalization gap (validation AUC 0.925 vs test AUC 0.677) confirms that the processing signal features learned during training do not transfer reliably to the test tracks. At the operating threshold, the model fires on approximately 48% of clean windows as false positives, yet the defective windows produce probabilities (0.49–0.50) that are indistinguishable from those of normal windows, providing no basis for confident defect localization. The agent framework, conditioned to require a localized cluster with meaningful elevation above the track mean, correctly abstains from asserting Defect under these conditions and produces 34 Uncertain decisions rather than compounding the model's unreliability with false confident outputs. In contrast, the monitoring agent achieves perfect defect detection (100% recall), confirming that acoustic signals provide a strong indicator of defect occurrence, although at the cost of moderate false positives. This performance is driven by the model's ability to capture temporally localized acoustic signatures of defects, which produce clear, high-contrast clusters that are easily distinguishable from the background signal.

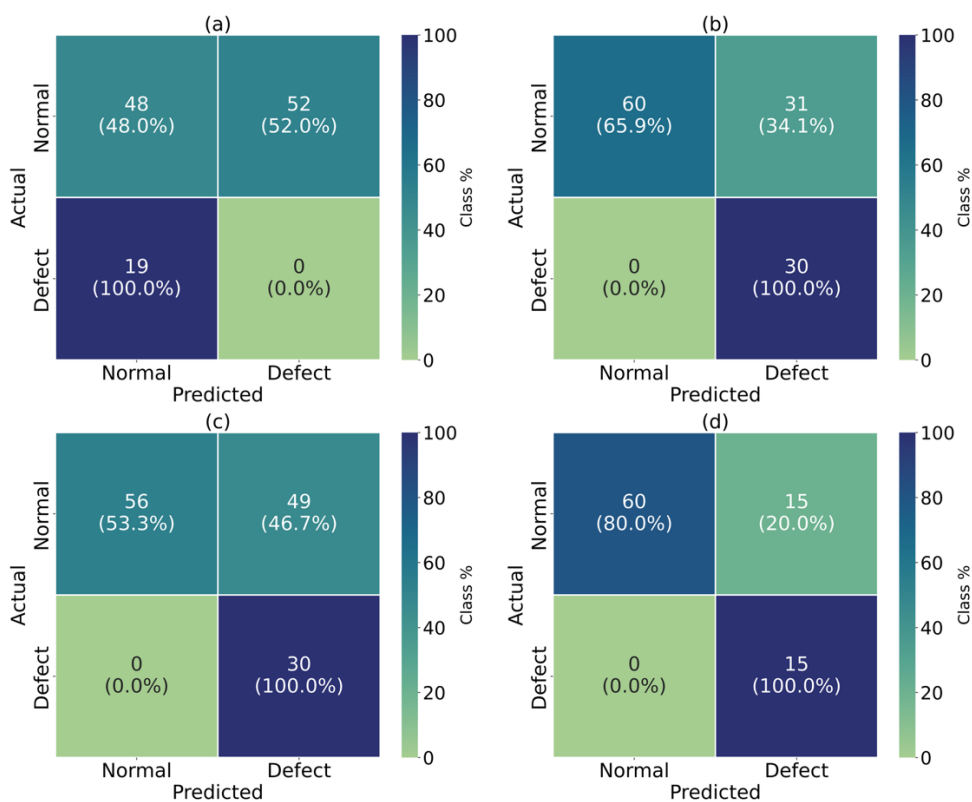

**Figure 3.** Confusion matrices for the four agent configurations on the held-out test set. Rows correspond to XCT ground truth labels (Normal, Defective); columns correspond to agent predictions. Cell values report the number of test tracks.

The combined single-agent model maintains perfect recall but does not significantly reduce false positives, indicating that naive fusion of modalities within a single reasoning step leads to suboptimal integration. This suggests that the weaker process signal modality introduces noise into the decision process rather than improving it. As a result, the model fails to effectively distinguish between meaningful defect patterns and noisy activations, leading to persistently high false positive rates despite the presence of a strong audio signal.

The multi-agent architecture achieves the best overall performance, retaining perfect defect detection while substantially reducing false positives. This improvement arises from structured orchestration, where the monitoring agent serves as the primary decision driver and the processing agent provides secondary validation. The orchestrator effectively filters unreliable signal contributions, leading to more precise and robust predictions.

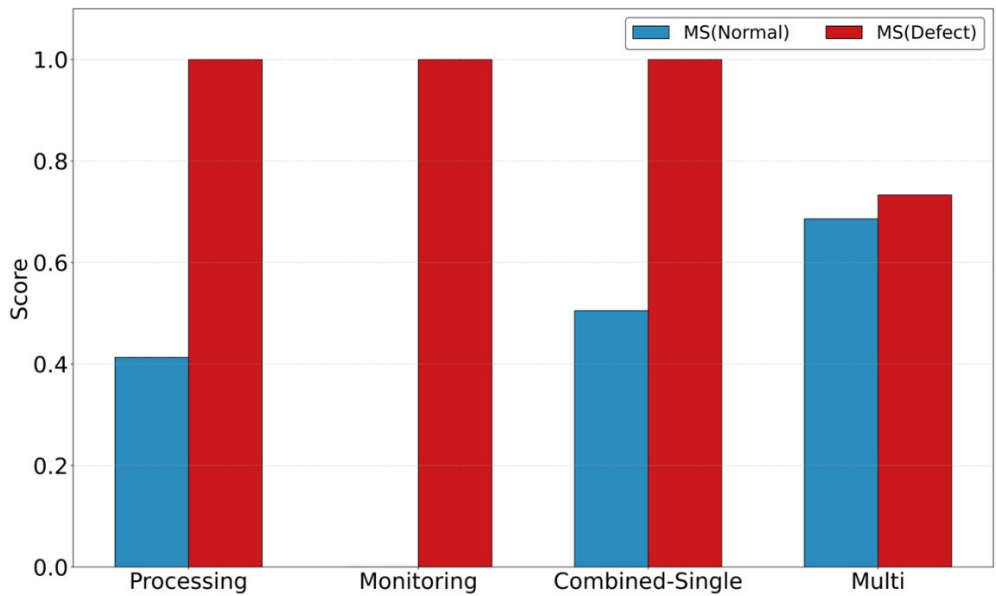

**Figure 4.** Class-wise monitoring score (MS) across different agent configurations. MS reflects the practical usefulness of monitoring recommendations.

**3.3 Monitoring utility**

The practical value of a process monitoring system depends on two complementary properties: its ability to flag every defective track for human review, and its ability to leave normal tracks undisturbed. A system that indiscriminately flags all instances to ensure complete defect coverage may incur excessive inspection overhead. To capture this trade-off, monitoring utility scores are reported separately for normal tracks MS(Normal) and defective tracks MS(Defect), where a higher MS(Normal) indicates fewer unnecessary alarms on clean tracks and a higher MS(Defect) indicates more complete defect coverage. **Figure 4** shows these scores across all four agent configurations.

Three of the four configurations (the processing agent, the monitoring agent, and the combined single agent) achieve MS(Defect) = 1.0, meaning every defective track was flagged in every run. However, this perfect defect coverage comes at a high cost. The monitoring agent produces MS(Normal) = 0.0, flagging every normal track as a potential defect and offering no discrimination between clean and defective tracks. The processing agent achieves MS(Normal) = 0.413, and the combined single agent MS(Normal) = 0.505, both indicating that roughly half of all normal tracks are incorrectly flagged for inspection. In a production setting, this level of false alarm rate would render the system impractical, as operators would be required to inspect most tracks regardless of their actual quality.

The multi-agent configuration achieves the most balanced monitoring profile. It reaches MS(Normal) = 0.686, the highest specificity across all configurations, meaning that approximately 69% of normal tracks are correctly cleared without human intervention. Its MS(Defect) = 0.733 reflects the fact that the remaining defect coverage is distributed between confirmed Defect decisions and Uncertain abstentions, all of which are flagged for monitoring. Crucially, no defective track receives a Normal decision in any run, meaning the multi-agent never silently misses a defect. The reduction in MS(Defect) relative to the other configurations, therefore, represents a deliberate redistribution from overconfident Defect calls to honest Uncertain flags, rather than a genuine loss of defect sensitivity.

### 3.4 Reasoning Quality

Reasoning quality was evaluated using an independent LLM judge scoring each agent's reasoning trace across three dimensions (FG, LC, and MSI) on a 1 to 5 Likert scale. The composite RQ score was computed as the arithmetic mean of the three dimensions as discussed in Section 2.6.3. **Figure 5** shows the per-dimension scores for all four configurations as radar charts, with the shaded region indicating the interquartile range across scored traces.

FG scores were consistently high across all configurations, ranging from 3.78 to 4.41, confirming that the agents faithfully reported the numerical tool outputs they received without fabricating sensor readings or confidence values. This is an important safety property for a process monitoring application, where an agent that hallucinates CNN outputs would produce unreliable decisions regardless of its reasoning strategy. The processing agent achieved the highest FG score of 4.41, reflecting its tendency to cite specific probability values and threshold comparisons explicitly in every trace.

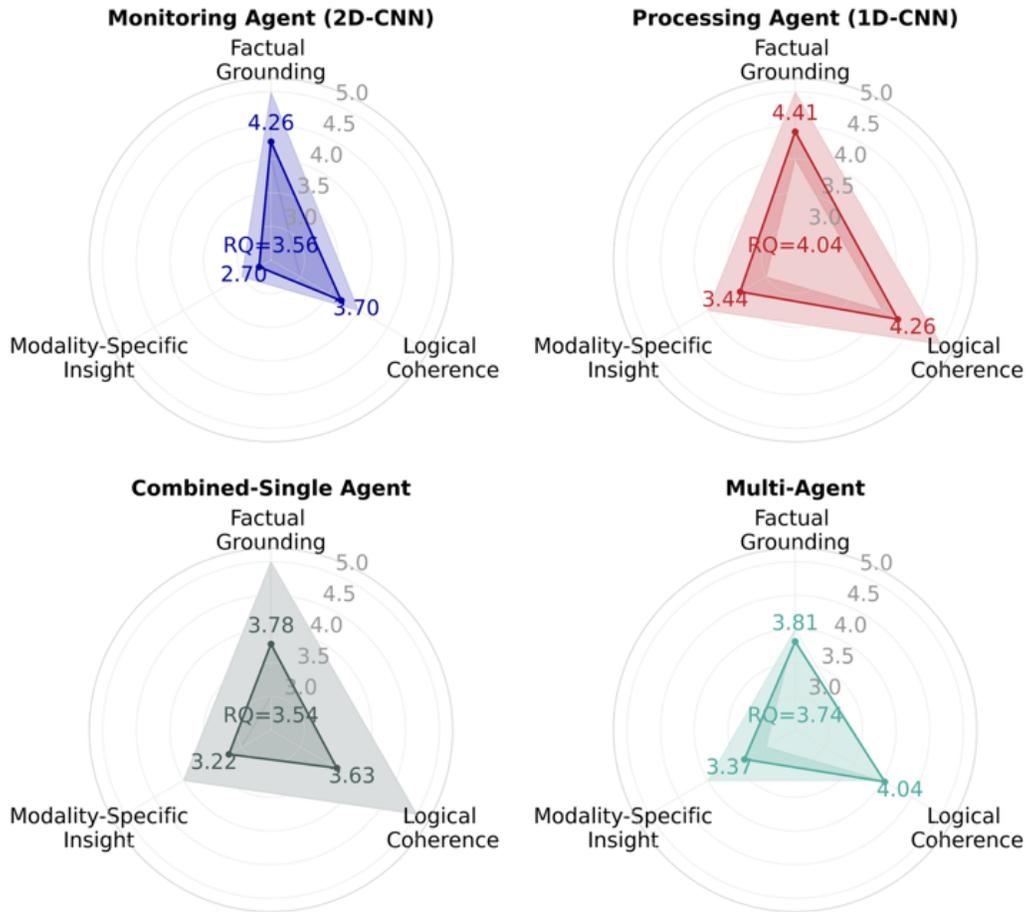

**Figure 5.** Per-configuration reasoning quality profiles across three evaluation dimensions, Factual Grounding (FG), Logical Coherence (LC), and Modality-Specific Insight (MSI), were scored by an independent LLM judge on a 1–5 Likert scale. Each panel shows the mean score per dimension for one agent configuration, with the shaded region indicating the interquartile range across scored traces. The composite Reasoning Quality score (RQ) is reported at the center of each panel. A larger and more symmetric polygon indicates more balanced and higher-quality reasoning across all three dimensions.

Another key observation is that LC varied more substantially across configurations. The multi-agent achieved the highest LC score of 4.04, indicating that its reasoning traces most consistently presented a clear and well-structured evidence-to-decision chain. The combined single-agent produced the lowest LC score of 3.63 with higher variance across tracks, suggesting that simultaneous reasoning over two modalities in a single inference step leads to less organized and less reliable argumentation than the specialist-then-synthesize approach of the multi-agent architecture.

MSI scores were lowest overall, ranging from 2.70 for the monitoring agent to 3.44 for the processing agent. The monitoring agent's low MSI score reflects that its traces rarely referenced the known limitations of the 2D-CNN model or the physical significance of audio spectrograms in the context of WAAM defect formation. The processing agent scored highest on MSI because its prompts explicitly required the agent to discuss the signal model's false positive rate, voltage insensitivity, and generalization gap in every trace, which is a systematic limitation-awareness that the judge consistently rewarded. The multi-agent MSI score of 3.37 reflects its intermediate position: the specialist sub-agents produce modality-aware reports, but the orchestrator's final synthesis tends toward a more concise integration that omits some of the per-modality detail present in the individual agent traces.

Composite RQ scores ranged from 3.54 for the combined single-agent to 4.04 for the processing agent. The processing agent's high RQ despite its poor detection accuracy illustrates an important distinction between reasoning quality and decision correctness. The judge rewards structured, limitation-aware reasoning regardless of whether the final decision is accurate. This decoupling of reasoning quality from detection performance highlights that RQ alone is not a sufficient criterion for selecting an agent configuration for deployment. For applications where both interpretability and accuracy are required, the multi-agent is the most appropriate choice, achieving the second highest RQ (3.74) alongside the best decision accuracy (91.6%) and the highest LC score (4.04), and produces reports that are both structurally sound and grounded in correct decisions. The combined single agent, by contrast, yields the lowest RQ (3.54), making its reasoning quality unpredictable and unsuitable for consistent evidence. Comparing all, the multi-agent architecture delivers the strongest balance across accuracy, LC, and reasoning consistency, and is recommended as the default configuration for interpretable in-situ WAAM process monitoring.

### 3.5 Agentic behavior: case studies

Two representative case studies are presented to illustrate the qualitative reasoning behavior of the agentic system: one normal track and one defective track. Each case study documents the tool outputs received by the agent(s), the resulting reasoning trace, and the final decision, and contrasts the single-agent and multi-agent responses for the same track.

### 3.5.1 Case study I: normal beads

In the first case study, a representative normal printed bead (layer 11, track 6) is used to assess the reasoning behavior of different agentic configurations. The individual processing and monitoring agents call the classification tool and produce consistent outputs indicating normal operation, with no significant peaks or localized clusters observed in their respective probability distributions. In the combined single-agent configuration (**Figure 6**), the agent directly integrates both modalities within a single reasoning step. The resulting explanation is concise and confirms the absence of defects based on the lack of threshold exceedance and uniform probability patterns. While the decision is correct, the reasoning remains aggregated, with limited distinction between the contributions of individual modalities.

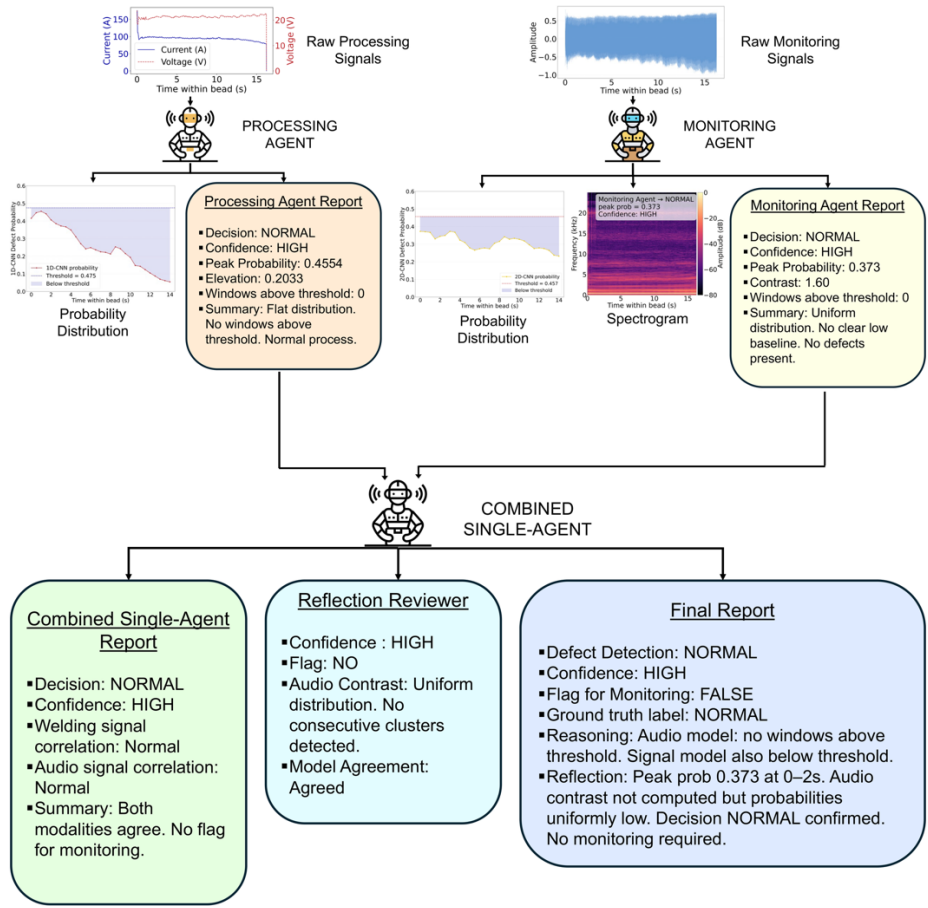

**Figure 6.** Normal bead monitoring using the combined single-agent configuration. The agent receives processing and monitoring signal inputs and integrates both modalities within a single reasoning step. The signal probability distribution and audio spectrogram show no significant peaks or localized clusters, resulting in a normal classification. The reasoning trace is concise and aggregated, confirming the absence of defects based on uniform probability patterns and lack of threshold exceedance.

In contrast, the multi-agent configuration (**Figure 7**) demonstrates a more structured reasoning process. The processing agent and monitoring agent independently analyze the processing and monitoring signals, each producing modality-specific summaries that consistently indicate normal behavior. The orchestrator agent then integrates these independent assessments, explicitly confirming agreement across modalities. The reflection stage further validates the absence of clusters and reinforces the consistency of the decision. This case highlights that, even under agreement conditions, the multi-agent framework provides greater transparency by preserving modality-specific reasoning traces. The decision is not only correct but also more interpretable, as the contribution of each modality is explicitly represented and validated through structured coordination.

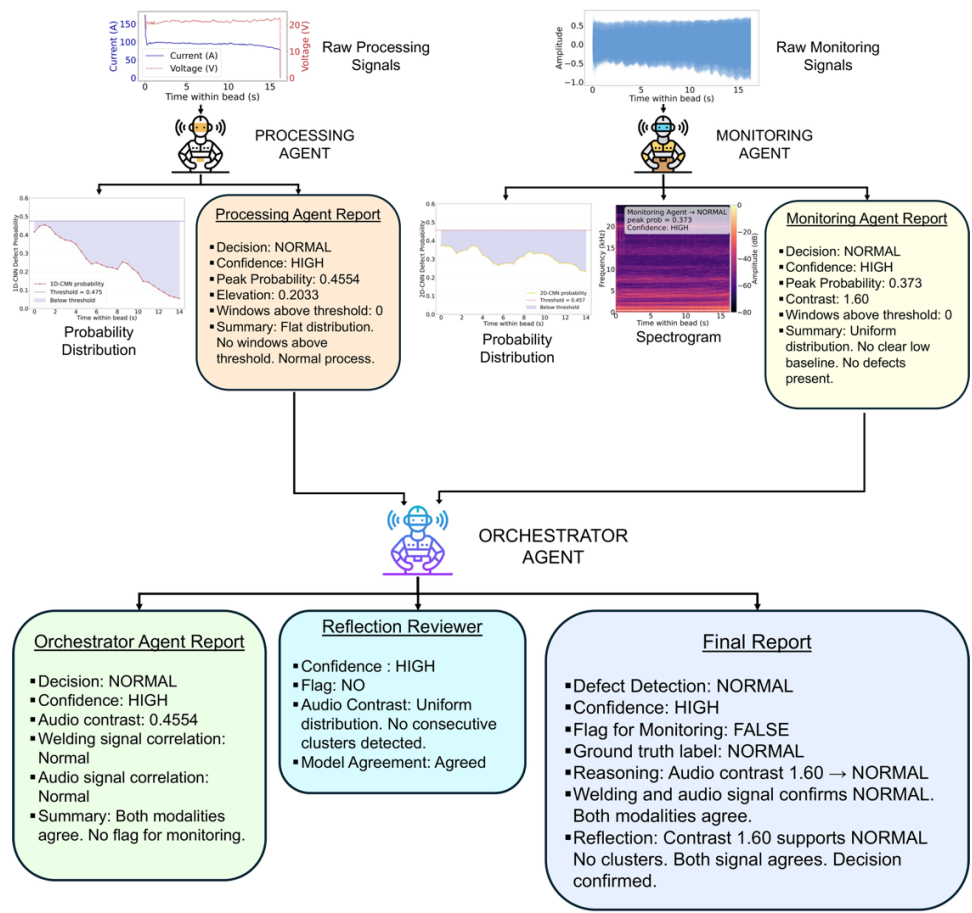

**Figure 7.** Normal bead monitoring using the multi-agent configuration. The processing agent and monitoring agent independently analyze welding signals and acoustic data, respectively, each producing modality-specific reports indicating normal behavior. The orchestrator agent integrates these outputs and confirms agreement across modalities, while the reflection stage validates the absence of localized clusters or contradictory evidence. The final decision is normal, with improved interpretability due to structured, modality-specific reasoning.

## 3.5.2 Case study II: defective beads

In this case study, a representative defective bead is used to assess the reasoning behavior of different agentic configurations. **Figures 8** and **9** present a representative defective track, illustrating system behavior under conflicting and ambiguous evidence. In this case, the processing agent indicates elevated activity with localized windows above threshold, while the monitoring agent shows high contrast but lacks a clearly defined consecutive cluster. As a result, the two modalities provide partially conflicting and weakly localized evidence of defect presence.

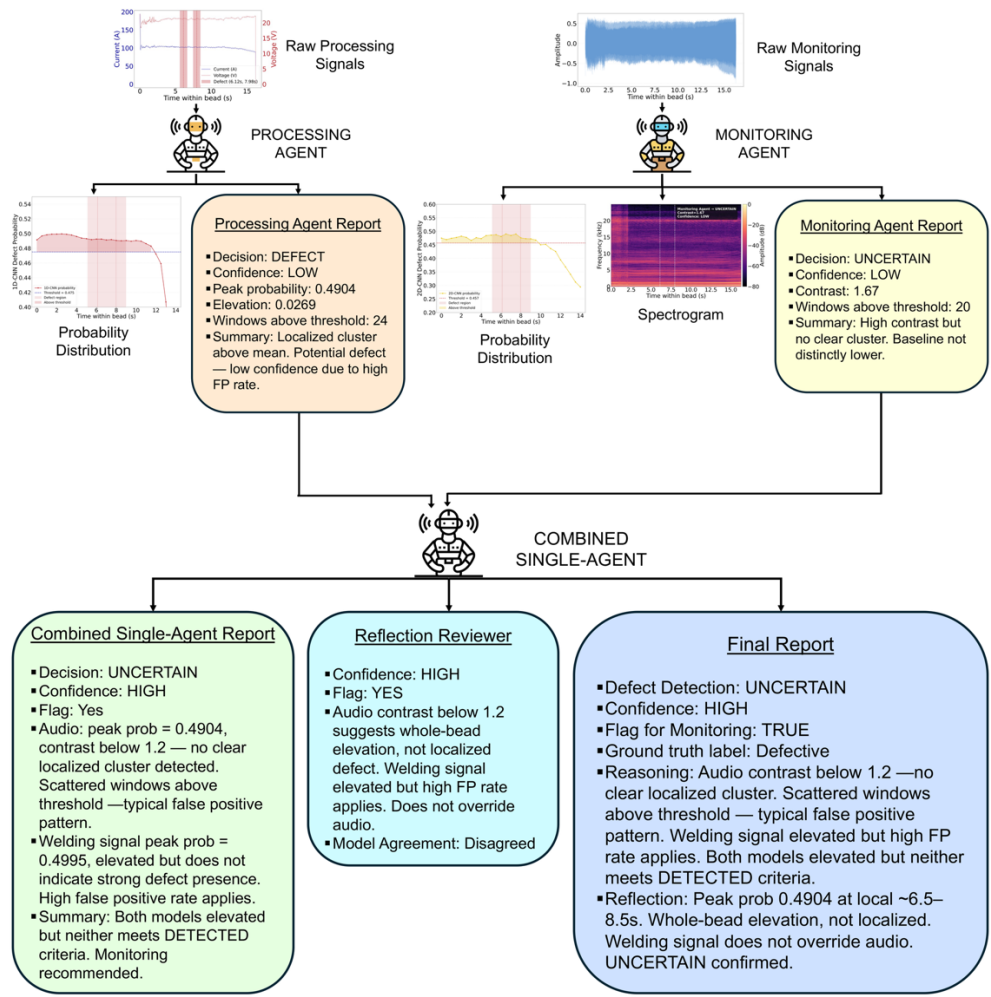

**Figure 8.** Defective bead monitoring using the combined single-agent configuration. The agent integrates processing and monitoring inputs within a single reasoning step, identifying elevated activity in both modalities. The signal probability distribution shows multiple windows above threshold, while the spectrogram exhibits increased contrast without a clearly localized cluster. The agent produces a high-confidence uncertain decision, reflecting overinterpretation of dispersed patterns and limited ability to distinguish between noise and true defect signatures.

In the combined single-agent configuration (**Figure 8**), both modalities are integrated within a single reasoning step. The agent identifies elevated responses in both modalities. Despite the lack of strong localization, the reasoning process produces a high-confidence uncertain decision, flagging the track for monitoring. However, the reasoning conflates dispersed signal elevations with defect evidence, leading to overconfident interpretation of ambiguous patterns.

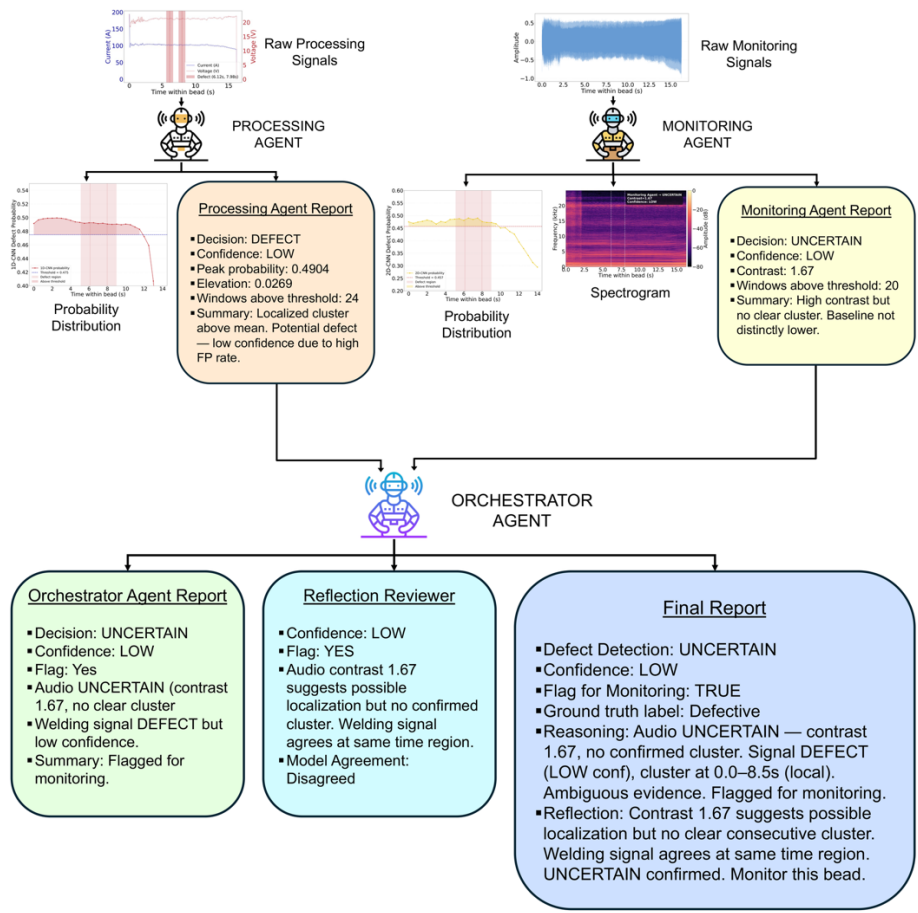

**Figure 9.** Defective bead monitoring using the multi-agent configuration. The processing and monitoring agents independently analyze processing and monitoring data, respectively, each identifying elevated activity but noting the absence of strong localization. The orchestrator agent integrates these modality-specific assessments and explicitly recognizes disagreement and ambiguity. The reflection stage confirms that neither modality provides sufficient evidence for a definitive defect classification. The final decision is uncertain with low confidence, demonstrating more cautious and reliable reasoning under ambiguous conditions.

In contrast, the multi-agent configuration (**Figure 9**) demonstrates more cautious and structured reasoning. The processing agent identifies elevated signal activity but explicitly notes the high false positive tendency and lack of strong localization. The monitoring agent detects increased contrast in the spectrogram but does not confirm a clear defect cluster. The orchestrator agent integrates these independent assessments and recognizes the disagreement and ambiguity between modalities. The reflection stage further reinforces that while both modalities show elevated activity, neither provides sufficient evidence to meet defect detection criteria. As a result, the multi-agent system produces an uncertain decision with low confidence, appropriately flagging the track for monitoring rather than committing to a potentially incorrect classification. This difference in confidence highlights a key advantage of the multi-agent framework: it avoids overconfident decisions under ambiguous conditions and instead reflects uncertainty in a controlled and interpretable manner.

## 4. Discussion

### 4.1 Relative performance of signal modalities

The monitoring agent substantially outperformed the processing agent under the conditions of this study, suggesting that audio signals carry richer porosity-relevant information than processing signals. This difference likely reflects what each sensor physically measures. Monitoring signals capture transient arc instability and spatter events that correlate directly with melt pool disturbances causing porosity, whereas processing signals reflect bulk arc impedance and are more sensitive to gradual process drift than to the localized sub-surface events identified by XCT. These findings are broadly consistent with prior WAAM monitoring literature, where acoustic features have shown higher defect classification accuracy than electrical features on similar datasets [34,110], though the ranking should not be over-generalized as the relative utility of each modality is likely process and material dependent.

### 4.2 Multi-agent advantage over single-agent

The multi-agent configuration outperformed the combined single-agent on both accuracy and reasoning quality, with the gain in precision being the primary driver, while both configurations maintained perfect recall. The performance improvement reflects the value of separating reasoning responsibilities in the combined single agent; one LLM call must simultaneously interpret two modalities and reach a decision, which can dilute modality-specific deliberation, whereas the multi-agent approach enforces a separation of concerns in which each specialist agent reasons about one modality before the orchestrator synthesizes a final verdict. The orchestrator's most distinctive contribution is adjudicating disagreement between the two specialist agents by weighing evidence according to each model's known reliability, making it less susceptible to being misled by the signal model's high false positive rate. The multi-agent also produced better calibrated confidence scores in the defective bead case study; the single agent returned UNCERTAIN with HIGH confidence while the multi-agent correctly returned UNCERTAIN with LOW confidence, a practically meaningful difference for prompting appropriate human review.

The primary cost of the multi-agent architecture is computational, requiring three LLM inference calls per track compared to one for the single agent. For the offline monitoring application described in this study, this overhead is acceptable, but for high-throughput lines requiring sub-second decisions, the single-agent configuration may be preferable. The two architectures, therefore, represent a clear engineering design choice. The multi-agent offers higher accuracy, better calibrated uncertainty, and more consistent reasoning, while the single agent offers lower latency at a modest cost to precision and reasoning stability.

### 4.3 Consistency, stochasticity, and deployment reliability

Despite the stochastic nature of LLM inference, the multi-agent configuration achieved 100% decision consistency across all 9 test tracks over 15 independent runs, while the combined single agent achieved 88.9% unanimity, with one borderline track producing mixed decisions. The high consistency observed is partly attributable to the use of temperature zero during inference, which makes the model always pick the same response for the same input. The occasional inconsistency seen in the single agent is not due to randomness in the model itself, but rather small differences in how the input is presented across runs, which can lead to slightly different outputs even when the underlying data is the same.

Unlike traditional machine learning classifiers, LLM agents are inherently stochastic, so measuring and reporting consistency is a necessary step before any industrial adoption. The results here suggest that the multi-agent framework is sufficiently consistent for a process monitoring application, and that majority voting across a small number of repeated runs is a viable strategy for borderline cases where a single run may not be reliable. Future work should investigate how consistency changes at higher inference temperatures, where more diverse reasoning may improve accuracy on ambiguous tracks at the cost of repeatability.

### 4.4 Interpretability, reasoning quality, and the role of Agentic AI

A key contribution beyond classification accuracy is interpretability. Unlike traditional CNN classifiers that produce only a confidence score, the agentic framework generates a structured, human-readable reasoning trace for every track documenting how the evidence was interpreted and why a decision was reached with direct practical value in AM qualification. FG scores were consistently high across all configurations, confirming that agents faithfully reported tool outputs without fabricating sensor readings, which is an important safety property in a production context. The multi-agent achieved the highest LC and MSI scores, with traces more consistently referencing physically meaningful phenomena specific to each modality. These findings extend recent LLM-for-AM work, including LLM-3D Print [96]and AMGPT [72], to in-situ defect monitoring, demonstrating that LLM agents can not only classify track quality but explain their classification in physically grounded terms. The reasoning quality evaluation carries two limitations worth noting: scores were assigned by an LLM judge rather than human experts, and the framework currently produces monitoring decisions without prescribing corrective actions, therefore, extending the orchestrator with deposition control tools represents the most impactful near-term direction for this work.

### 4.5 Limitations

Additional limitations specific to experimental design and system implementation are discussed below; limitations arising from the evaluation methodology and computational trade-offs have been noted in the relevant subsections above.

Several limitations bound the scope of the present findings. The dataset contains approximately 90 tracks from a single WAAM machine using the same wire material and shielding gas throughout, which means the results may not hold for different materials, machine setups, or deposition conditions. Additionally, the results depend on the specific version of the language model used, and future model updates in this area could change how the agents behave without any changes to the code. Testing how sensitive the results are to different prompt wordings was also not carried out and would be a useful step before any real-world deployment.

### 5. Conclusion

This paper presented an agentic AI framework for in-situ process monitoring of WAAM, integrating a 1D-CNN for processing signals and a 2D-CNN for monitoring signals classification as callable tools within a LangGraph-based multi-agent system powered by an LLM brain. Four configurations were evaluated: a processing agent, a monitoring agent, a combined single agent, and a multi-agent system comprising specialist agents coordinated by an orchestrator.

The monitoring agent outperformed the processing agent, confirming that acoustic signals carry richer porosity-discriminative information under the tested conditions. The multi-agent configuration achieved the highest overall performance, representing a meaningful improvement over the combined single agent in both precision and specificity, while maintaining perfect defect recall and producing the most balanced monitoring utility across normal and defective tracks. The multi-agent configuration also achieved the most balanced monitoring utility, correctly clearing 68.6% of normal tracks without unnecessary inspection while maintaining full defect coverage through a combination of confirmed detections and conservative uncertain flags. This shows that the framework avoids the all-flag behavior exhibited by the single modality agentic configurations. Reasoning quality evaluations showed that the multi-agent produced the most logically coherent traces with the highest modality-specific insight, demonstrating the value of specialist deliberation at the sub-agent level.

Beyond classification accuracy, the framework generates structured, human-readable reasoning traces for every monitoring decision, providing a form of process documentation that conventional classifiers cannot offer and that may serve as evidence in AM part qualification workflows. The proposed framework demonstrates the potential of a LangGraph-orchestrated multi-agent AI system for in situ WAAM process monitoring and provides a foundation for autonomous, explainable, and certifiable quality assurance in metal additive manufacturing. Future work will focus on extending the framework to multi-material and multi-process configurations, integrating corrective action tools to close the monitoring control loop, and validating reasoning quality against human expert annotations.

**Acknowledgements**

S.M. acknowledges support from the Washington State University start-up fund for this research. S.M. and P.H. also acknowledge Oak Ridge National Laboratory for providing the public dataset used in this study.

# Appendix

## A. Dataset and model training details

The dataset comprises 90 tracks in total across 15 layers and 6 tracks per layer, of which 18 are defective, and 72 are clean as determined by post-deposition XCT scanning. A stratified track-level 80/10/10 split was applied with a fixed random seed (42), ensuring no window-level leakage between sets. The training set contains 72 tracks (14 defective, 58 clean), the validation set 9 tracks (2 defective, 7 clean), and the test set 9 tracks (2 defective, 7 clean). The nine test tracks used for all evaluations reported in this paper are listed in **Table A1**. Both CNN models share the identical track split so that all comparisons are made on the same held-out data.

The $F_1$-optimal operating threshold for each model was selected on the training set predictions and embedded in the model checkpoint. Both models were configured for zero missed detections at the operating threshold, accepting a higher false positive rate in exchange for perfect recall. This deliberate design choice motivates the multimodal reasoning framework described in the main paper.

**Table A1**. Test track list with ground truth labels.

| Layer | Track | Ground Truth |
|---|---|---|
| 1 | 5 | Normal |
| 3 | 6 | Normal |
| 5 | 3 | Defect |
| 7 | 6 | Normal |
| 8 | 2 | Normal |
| 8 | 3 | Defect |
| 11 | 6 | Normal |
| 12 | 6 | Normal |
| 15 | 3 | Normal |

## B. Agent system prompt

The agentic framework uses eight distinct system prompts across all four configurations. Each prompt follows a four-step structure: a persona definition establishing the agent's role and modality scope, a task description specifying the tool pipeline to execute, a model context section providing the CNN's performance characteristics and decision rules, and a structured output format that the agent must follow exactly.

The processing and monitoring only configurations each use two prompts: a main system prompt and a reflection prompt. The main prompts instruct the respective agent to run its full tool pipeline independently and produce a final detection decision. The combined single-agent uses a similar two-prompt structure but instructs the agent to run both pipelines sequentially and synthesize the evidence into a single decision, treating the audio model as the primary source.

The multi-agent configuration uses four prompts. The monitoring agent prompt and processing agent prompt differ from their single-modality counterparts, as rather than making a final decision, each specialist is instructed to produce a structured modality report in a fixed format, which is passed to the orchestrator as context. The orchestrator prompt receives both reports and explicitly instructs the model on how to weight the two sources based on each model's known reliability, favoring the monitoring agent while treating the processing agent report as secondary context. The reflection prompt used by the quality assurance reviewer is shared between the combined single agent and multi-agent configurations, because the review task of checking the logical consistency and confidence calibration of the final decision is the same in both cases. All nine prompts are available in the code repository accompanying this paper and described in **Appendix B**.

## C. LLM judge scoring rubric

The full scoring rubric provided to the LLM judge is reproduced below for reproducibility. The judge was instructed to respond with a JSON object containing integer scores (1–5) and a one-sentence justification for each dimension, with no additional commentary. Inference temperature was set to zero. The complete judge system prompt is available in the code repository.

Table A2. Factual Grounding (FG) scoring criteria

| Score | Scoring Criteria |
|---|---|
| 5 | Perfect fidelity to all reported tool outputs. All cited numerical values match what the tools returned. No hallucinated information. |
| 4 | Minor omissions of secondary evidence, but all key values cited correctly. No hallucination. |
| 3 | Some numbers are cited, but key evidence is missing or imprecisely stated. |
| 2 | Significant omissions or inaccuracies. Some values appear fabricated or inconsistent with tool outputs. |
| 1 | Substantial misrepresentation of tool outputs. Key values are wrong, missing, or clearly hallucinated. |

**Table A3.** Logical Coherence (LC) scoring criteria:

| Score | Scoring Criteria |
|---|---|
| 5 | Clear, well-organized argument with explicit causal links. Decision follows necessarily from the stated evidence. No contradictions. |
| 4 | Mostly logical with minor gaps. Decision is defensible from the stated evidence. |
| 3 | Reasoning present but with notable gaps or weak causal links. |
| 2 | Partially contradictory or poorly structured. Significant logical gaps. |
| 1 | Disjointed or self-contradictory. Conclusions do not follow from stated premises. |

**Table A4.** Modality-Specific Insight (MSI) scoring criteria

| Score | Scoring Criteria |
|---|---|
| 5 | Nuanced interpretation distinguishing between modalities with correct reference to model-specific limitations such as the signal model's false positive rate, the audio model's compressed probability range, and the contrast criterion for localization. |
| 4 | Good modality awareness with correct reference to at least one modality-specific limitation. |
| 3 | Some domain awareness but reasoning could apply to any two-modality classifier. |
| 2 | Minimal modality distinction. Tool outputs treated largely interchangeably. |
| 1 | Generic reasoning with no distinction between modalities and no reference to known model limitations. |

**Data Availability Statement**

The code developed in this study, including all agent configurations, system prompts, and evaluation scripts, is publicly available at https://github.com/Pallock-35/Agentic_AI_in_Process_Monitoring_WAAM. The WAAM process monitoring dataset used for model training and evaluation is publicly available and is cited in the references section of this paper.

**Declaration of generative AI and AI-assisted technologies in the manuscript preparation process**

During the preparation of this work the authors used Claude AI opus 4.6 model in order to assist with language editing only. After using this tool/service, the authors reviewed and edited the content as needed and take full responsibility for the content of the published article.